\documentclass[10pt,twocolumn,letterpaper]{article}

\usepackage[pagenumbers]{cvpr} 

\definecolor{cvprblue}{rgb}{0.21,0.49,0.74}
\usepackage[pagebackref,breaklinks,colorlinks,allcolors=cvprblue]{hyperref}
\usepackage[ruled,linesnumbered,algo2e]{algorithm2e}
\usepackage{multirow}
\usepackage{float}
\usepackage{tabularx}
\usepackage{graphicx}

\def\paperID{0020} 
\def\confName{CVPR}
\def\confYear{2025}

\title{Dual Precision Quantization for Efficient and Accurate Deep Neural Networks Inference}

\author{
Tomer Gafni\thanks{Corresponding author} \\ \and Asaf Karnieli \\ Intel, Israel \and Yair Hanani \\
\and {\tt\small \{tomer.gafni,asaf.karnieli,yair.hanani\}@intel.com}
}

\begin{document}
\maketitle
\begin{abstract}
Deep neural networks have achieved state-of-the-art results in a wide range of applications, from natural language processing and computer vision to speech recognition.
However, as tasks become increasingly complex, model sizes continue to grow, posing challenges in latency and memory efficiency. To meet these constraints, post-training quantization has emerged as a promising solution. In this paper, we propose a novel hardware-efficient quantization and inference scheme that exploits hardware advantages with minimal accuracy degradation. Specifically, we introduce a \textit{W4A8} scheme, where weights are quantized and stored using 4-bit integer precision, and inference computations are performed using 8-bit floating-point arithmetic, demonstrating significant speedups and improved memory utilization compared to 16-bit operations, applicable on various modern accelerators. To mitigate accuracy loss, we develop a novel quantization algorithm, dubbed Dual Precision Quantization (DPQ), that leverages the unique structure of our scheme without introducing additional inference overhead. Experimental results demonstrate improved performance (i.e., increased throughput) while maintaining tolerable accuracy degradation relative to the full-precision model.
The source code is publicly available at the Intel Neural Compressor  \href{https://github.com/intel/neural-compressor}{repository}.
\end{abstract}    
\section{Introduction}
\label{sec:introduction}

Deep neural networks (DNNs) have demonstrated remarkable capabilities across a broad spectrum of tasks, having a significant impact in numerous fields. In recent years, model sizes have grown significantly to better capture complex data structures, accommodate longer context lengths, and enhance overall model strength. While these large-scale models exhibit outstanding functionality, their immense size poses significant challenges for deployment, including high computational costs and memory demands. To address these issues and enable efficient inference, quantization techniques are essential.

Quantization methods can be broadly categorized into two approaches: quantization-aware training (QAT) \cite{QAT1, QAT2, QAT3, QAT4} and post-training quantization (PTQ)  \cite{nagel2019data, hubara2020improving, hubara2021accurate, liu2021post, nahshan2021loss, ZeroQuant}. The former integrates quantization into the training process, typically requiring extensive retraining and/or fine-tuning while employing approximate differentiation techniques for the rounding operation. In contrast, PTQ quantizes a pre-trained model with significantly fewer resources, such as compute and data, and over a much shorter process. PTQ is particularly appealing for large-scale models, where full retraining or even fine-tuning is prohibitively expensive. PTQ has been successfully implemented for both language models \cite{AdaRound, gptq, awq} and vision models \cite{VISION1, VISION2, CVPR1, CVPR4}.
In this paper, we focus on PTQ as a practical solution for efficient deployment of massive models.

PTQ produces a smaller and faster model. Reducing the size of the model allows us to use less memory on hardware accelerators such as Nvidia GPUs and Intel HPUs and/or increase the maximum batch size. A smaller model also reduces the bandwidth required to move weights from memory, which is particularly important at low batch sizes and in auto-regressive decoding scenarios. In addition, if we quantize the activations, we can take advantage of hardware support for fast low-precision operations, where the matrix-multiplication computation can be performed 2x faster than 16-bit floating point arithmetic \cite{fp8formates, lee2025faster}. Therefore, when considering PTQ, it is crucial to understand the different regimes of inference performance, which broadly fall into two categories: compute-bound inference, where performance is limited by computational throughput, and memory-bound inference, where the bottleneck in performance is the rate at which data is transferred from memory to processing cores. Memory-bound scenarios are especially typical at low batch size, where there are a small number of operations per byte of the loaded model.
In memory-bound cases, performance gains are typically achieved by reducing memory traffic rather than increasing compute power, as hardware compute units are often underutilized while waiting for data to be fetched from memory.
Taking this into account, two primary approaches are commonly used to accelerate and optimize DNN inference, as detailed next.

\textit{FP8 quantization}: Recent AI accelerators, including NVIDIA H100 \& H200, AMD MI300X, and Intel Gaudi 2 \& 3, supports the FP8 numerical format \cite{fp8formates, fp8_exponent, lee2025faster, kim2025investigation}. FP8 offers two key advantages over 16-bit floating-point (FP) formats: (i) cutting memory requirements in half, leading to better memory utilization and improved bandwidth efficiency; and (ii) reducing computational complexity, often doubling throughput compared to 16-bit FP operations \cite{FP8_versus_INT8, qserve, super, smoothquant}.

\textit{INT4 quantization}: For even greater compression, INT4 quantization is a leading approach that balances high compression with acceptable accuracy degradation. Storing weights in only 4 bits significantly reduces memory consumption, allowing larger models to fit on fewer machines and to increase the batch size, while improving inference speed. The low-precision representation reduces the latency associated with transferring weights from memory to processing cores, further accelerating model inference.
Various works considered effective algorithms, including AWQ \cite{awq}, GPTQ \cite{gptq}, QuaRot \cite{quarot}, SqueezeLLm \cite{squeezellm}, AdaRound \cite{AdaRound}, APQ-Vit \cite{ding2022towards}, NoisyQuant \cite{Noisyquant} and more \cite{AutoRound, OmniQuant, HQQ, PTQ_CVPR, Pdquant, liu2021post}.
Some algorithms related to this work uses information from second-order derivatives of the error function to perform network pruning and quantization \cite{optimalbrainsurgery, OBQ, yuan2022ptq4vit, gptq}.

In this work, we propose a hybrid W4A8 scheme, that both stores weights in 4-bit  precision (W4) accounting for the memory-bound regime, and performs the arithmetic in FP8 (A8), accounting for the compute-bound regime.
Recently, \cite{qserve} and \cite{qqq} proposed a W4A8 quantization scheme similar to the one presented in this paper. However, their approach quantizes activations to 8-bit integers (INT8), whereas our method employs FP8 for both activations and computations.

FP8 can be advantageous over INT8 in terms of model accuracy, as INT8 represents a uniform distribution of values, making it most effective when model weights and activations also follow a uniform distribution. However, this is rarely the case, as empirical studies show that pretrained neural network weights typically exhibit a zero-centered, normal-like distribution \cite{qlora}. In contrast, FP8, with its non-uniform distribution and higher dynamic range, better accommodates such data distributions while also providing higher precision for small values, and better captures outliers.
This advantage becomes even more significant when quantizing activations as well, since their distributions tend to be less uniform and contain more outliers compared to weight distributions.

Our main contributions are as follows. \\
    1)  We propose a novel quantization and inference scheme that effectively accelerates inference in both the compute-bound and memory-bound regimes. 
    Our proposed W4A8 scheme significantly reduces memory consumption, enabling larger batch sizes for faster evaluation and generation. Additionally, all matrix multiplications are performed in FP8, leveraging the capabilities of modern AI accelerators to enhance both speed and efficiency. \\
    2) To preserve accuracy, we introduce a dedicated quantization algorithm tailored for this scheme: the Dual Precision Quantization (DPQ). DPQ is designed to minimize quantization errors, allowing us to maintain most of the accuracy of the original model on various tasks, without introducing any inference overhead. \\ 
    3) Group-Aware Reordering (GAR): To further mitigate accuracy degradation, we introduce a weight quantization strategy that prioritizes important weights without inference overhead. Notably, GAR is not limited to our specific setting; it is applicable to other quantization schemes, making it a versatile technique for improving the accuracy of quantized models without introducing additional overhead. Ablation studies further confirm its effectiveness.

Finally, we conduct experiments on both language-vision and large language models across a range of tasks and sizes. Specifically, we evaluate our scheme on Llama-2 and Llama-3 models (7B to 70B) as well as the Qwen2-VL family (2B to 72B), comparing it against existing algorithms and alternative precision strategies. The results demonstrate significant throughput gains with minimal accuracy degradation compared to higher-precision schemes, while also achieving better accuracy than existing W4A8 schemes, validating the effectiveness of our approach.

\section{Background}
\label{sec:Background}

In this section, we introduce the fundamentals of DNN quantization, establish the relevant notation for our scheme, and discuss the key challenges in PTQ.

\subsection{Integer Quantization} \label{ssec:INT}
Integer quantization maps high-precision numbers to discrete levels. The process can be formulated as:

\begin{equation} \label{eq:int_quantization}
\begin{aligned}
\boldsymbol{X}_q^{\text{INT}} &= \left\lceil{\frac{\boldsymbol{X}}{s_I} + z_I} \right\rfloor, \\ 
s_I = \frac{\boldsymbol{X}_\text{max} - \boldsymbol{X}_\text{min}}{q_{\text{max}} - q_{\text{min}}}, & \quad 
z_I = \left\lceil{q_{\text{min}} - \frac{\boldsymbol{X}_\text{min}}{s_I}} \right\rfloor,
\end{aligned}
\end{equation}
where $\lceil{\cdot}\rfloor$ denotes the rounding-to-nearest-integer operator, $\boldsymbol{X}$ represents the high-precision FP tensor, and $\boldsymbol{X}_q^{\text{INT}}$ is its $n$-bit quantized integer counterpart. The parameters $s_I$ and $z_I$ denote the scaling factor and zero-point, respectively, where the subscript $I$ indicates the integer format. 
This is known as \textit{asymmetric} quantization, where $\boldsymbol{X}_{\text{max}} = \max(\boldsymbol{X})$, $\boldsymbol{X}_{\text{min}} = \min(\boldsymbol{X})$, and $q_{\text{max}} - q_{\text{min}} = 2^n - 1$. \cref{eq:int_quantization} can be further simplified to \textit{symmetric} quantization, where $z_I =0$, $\boldsymbol{X}_{\text{max}} = -\boldsymbol{X}_{\text{min}} = \max|\boldsymbol{X}|$, and $q_{\text{max}} - q_{\text{min}} = 2^n - 2$. 

The dequantized tensor is represented as:
\begin{equation} \label{eq:int_dequantization}
\hat{\boldsymbol{X}} = (\boldsymbol{X}_q^{\text{INT}} - z_I) \cdot s_I.
\end{equation}

In this paper, weights are stored in 4-bit precision using asymmetric integer quantization.

\subsection{FP8 Quantization} \label{ssec:FP8}
In this section, we describe the FP8 quantization scheme used in this paper, focusing on the E4M3 data type (4-bit exponent and 3-bit mantissa). This format is widely adopted for inference, as higher precision is generally more important than a large dynamic range, which is more relevant during training (particularly in the backward pass) where the E5M2 format is commonly used.
Similarly to integer quantization, FP8 quantization applies a scaling factor to map a high-precision tensor into the FP8 dynamic range, followed by a rounding operation to the nearest representable value in the FP8 grid. We employ symmetric quantization. The process can be mathematically expressed as:

\begin{equation} \label{eq:FP8_quantization}
    \boldsymbol{X}_{q}^{\text{FP}} = \text{nearest}_{\mathcal{Z}} (\frac{\boldsymbol{X}}{s_F} ), s_F = \frac{\max(|\boldsymbol{X}|)}{Z_{\max}}.
\end{equation}
Here, $\mathcal{Z}$ represents the E4M3 grid, and $Z_{\max}$ is its maximum value. The scaling factor $s_F$ is used for FP quantization (with the subscript $F$ indicating the float type). The tensor $\boldsymbol{X}_{q}^{\text{FP}}$ denotes the $q$-bit FP quantized representation (FP8 in this paper). The corresponding dequantized tensor is given by:
\begin{equation} \label{FP_dequantization}
\hat{\boldsymbol{X}} = \boldsymbol{X}^{\text{FP}}_q \cdot s_{\text{F}}.    
\end{equation}

When implementing quantization algorithms, it is essential to consider the specific characteristics of the target hardware. For example, a key optimization in Intel Gaudi 2 and 3 accelerators is that when both input tensors (i.e., weights and activations) use scaling factors that are powers of two, scaling can be efficiently implemented by adjusting the exponent bias instead of multiplying individual elements \cite{lee2025faster}. This reduces computational overhead and improves FP8 throughput by several percentage points. 
To take advantage of this optimization, we incorporate the power-of-two scaling factors into our quantization scheme. For an in-depth analysis of using a power-of-two scaling factor, please refer to \cite{lee2025faster}.

In this paper, integer quantization is applied only to weight tensors for storage, while FP quantization is used for both weight and activation tensors at computation time.

\subsection{PTQ - Main Challenges} \label{challenges}
The quantization process in DNNs poses several fundamental challenges that must be carefully addressed when implementing PTQ algorithms. Below, we outline two key challenges relevant to our framework.

A primary issue is the presence of outliers, which force the scaling factor to accommodate an extended dynamic range. This results in the compression of most tensor values into a narrower range, increasing quantization step size and leading to higher reconstruction errors, which can significantly degrade model accuracy. To mitigate this, various techniques have been proposed, including clipping extreme values, per-channel or per-group quantization, and rotation methods that better capture the true distribution of values \cite{quarot, SpinQuant}

Another key challenge in PTQ is that not all weights and activations contribute equally to the network’s final output. As a result, quantization errors in highly influential components can have a disproportionate impact on overall model accuracy.  To address this, several strategies have been explored, including reordering the weight matrix based on activation magnitudes to prioritize more critical elements \cite{gptq}, scaling weights and activations in a way that preserves the most important values while reducing errors in less significant ones \cite{awq}, and restoring key scalar values that have an outsized influence \cite{massive, super}.
These algorithms help maintain accuracy in quantized models by focusing on the most significant elements, ensuring a more robust and efficient quantization process.
\section{DPQ: Dual Precision Quantization for W4A8} 
\label{sec:Method}
We now introduce our Dual Precision Quantization (DPQ) algorithm. DPQ follows a structured flow: first, high-precision weights are quantized to INT4 offline. During inference, the 4-bit weights are loaded and dequantized to FP8. Simultaneously, activations are quantized from brain float (BF) 16 to FP8 to leverage efficient FP8 matrix multiplication, with the output in high-precision (BF16).
To ensure clarity in the following formulation, we explicitly denote bit precision using subscripts to eliminate any ambiguity. 

\subsection{Compensating for Quantization Errors in W4A8 Flow} \label{ssec:compensating}
Note that our proposed target inference scheme involves two levels of weight quantization: 8-bit FP precision and 4-bit integer precision. 
Each quantization step introduces an error due to the rounding operator, which can degrade the accuracy of the model.

To mitigate this error, we take inspiration from the GPTQ algorithm \cite{gptq} and build upon the Optimal Brain Quantization (OBQ) method \cite{OBQ} for layer-wise quantization. Specifically, our objective is to find a 4-bit weight matrix, such that its dequantized 16-bit matrix,  $\hat{\boldsymbol{W}}_{16}$, minimizes the squared error with respect to the full-precision BF16 layer output:
\begin{equation} \label{The GPTQ_algorithm}
    \text{argmin}_{\hat{\boldsymbol{W}}_{16}} ||\boldsymbol{W}_{16}\boldsymbol{X}_{16} - \hat{\boldsymbol{W}}_{16}\boldsymbol{X}_{16}||^2_2,
\end{equation}
where $\hat{\boldsymbol{W}}_{16}$ incorporates the two-step dequantization process: from INT4 to FP8, and then from FP8 to BF16, and will be formally defined later in this section. This ensures that the error in \cref{The GPTQ_algorithm} accounts for the cumulative error introduced by both quantization levels, as this is the error that will be encountered during inference.

As in GPTQ, we quantize one weight at a time for each row in $\boldsymbol{W}_{16}$. 
To mitigate the quantization error, we distribute the resulting error among subsequent weights in the row (i.e., from left to right). Each weight update is scaled by a factor determined by the Hessian matrix $\boldsymbol{H}$, which accounts for the importance of each weight.

We now provide a detailed description of the DPQ algorithm. First, we compute a scaling factor to convert the weights from BF16 to FP8, either per tensor or per channel. For clarity, we assume the scale is computed per tensor, denoted by $s_{w, 16 \rightarrow 8}$:
\begin{equation} \label{scale_bf16_to_fp8}
    s_{w, 16 \rightarrow 8} = \frac{\max (|\boldsymbol{W}_{16}|)}{Z_{\max}}.
\end{equation}
Next, consider the first weight in a given row of the weight matrix, denoted by $w_{16}$. The procedure is applied independently to all weights in the first column, but we describe it for a single row for clarity. We quantize $w_{16}$ to FP8 using symmetric quantization:
\begin{equation} \label{quantized w_8}
    w_8^{\text{FP}} = \text{nearest}_{\mathcal{Z}} (\frac{w_{16}}{s_{w, 16 \rightarrow 8}}).
\end{equation}
We then compute the per-group scale and zero-point for converting FP8 to INT4. These values are determined, for a given group $g$ (which is represented in FP8 precision), once at the beginning of the group quantization process:
\begin{equation}  \label{scale_FP8_to_INT4}
\begin{aligned}    
         s_{8 \rightarrow 4}^g =& \frac{\max(\boldsymbol{W}_8^g) - \min(\boldsymbol{W}_8^g)}{q_{\max}}, \\ z_{8 \rightarrow 4 }^g =& \lceil{-\frac{\min(\boldsymbol{W}_8^g)}{s_{8 \rightarrow 4}^g}\rfloor},
\end{aligned}
\end{equation} 
where $q_{\max}=15$ for INT4 unsigned quantization as done in this paper.
Using the computed scale and zero-point, we quantize the weight to INT4 using asymmetric integer quantization:
\begin{equation} \label{quantized w_4}   
    w_4^{INT} = \lceil{\frac{w_8^{FP}}{s_{8 \rightarrow 4 }^g}  +  z_{8 \rightarrow 4}^g \rfloor}.   
\end{equation}
The weight $w_4^{INT}$ is the weight which will be stored and used in inference. Following OBQ, we now update the remaining weights in the row (which were not quantized yet), to compensate for the quantization error. This process includes few steps.
First, we dequantize the 4-bit weight back to FP8:
\begin{equation} \label{dequantized w_8}
    \hat{w}_8^{FP} = (w_4^{INT} - z_{8 \rightarrow 4}^g) \cdot s_{8 \rightarrow 4}^g.
\end{equation}
Then, we dequantize $\hat{w}_8^{FP}$ to BF16:
\begin{equation} \label{dequantized w_16}
    \hat{w}_{16} = \hat{w}_8^{FP} \cdot s_{w, 16 \rightarrow 8}.
\end{equation}
We then compute the quantization error and scale it by the inverse of the corresponding diagonal element of the Hessian, denoted by $[\boldsymbol{H}^{-1}]_{qq}$, where $q$ is the index of the currently quantized weight.

Next, let $F$ denote the set of remaining full-precision weights, i.e., the weights positioned to the right of the currently quantized weight.
The optimal update for a weight $w_v \in F$, denoted by $\delta_{F}$, is obtained by multiplying the scaled quantization error with the corresponding off-diagonal Hessian element $(\boldsymbol{H}_F^{-1})_{q,v}$, which captures the interaction between the quantized weight and the weight being updated: :
\begin{equation} \label{algorithm_update}
    w_v \leftarrow w_v + \delta_{F}, \quad    
    \delta_{F} = -\frac{w_{16} - \hat{w}_{16}}{[\boldsymbol{H}_F^{-1}]_{qq}} \cdot (\boldsymbol{H}_F^{-1})_{qv}.   
\end{equation}
The Hessian matrix characterizes the sensitivity of the weight to small perturbations, capturing how changes in the weight affect the loss function. The derivation of the optimal update in \cref{algorithm_update} can be found in \cite{OBQ}.

This procedure is repeated iteratively for all remaining weights in the row. 
A pseudocode for the proposed DPQ algorithm is provided in \cref{pseudo}.

Recall that our scheme also includes activation quantization. Specifically, the activation input $\boldsymbol{X}_{16}$, which serves as the input to the linear layer, is quantized to FP8 using symmetric quantization:
\begin{equation} \label{activation fp8 quantization}
\boldsymbol{X}_8 = \text{nearest}_{\mathcal{Z}} (\frac{\boldsymbol{X}_{16}}{s_{x,16 \rightarrow 8}}), s_{x,16 \rightarrow 8} = \frac{\max (|\boldsymbol{X}_{16}|)}{Z_{\max}}.
\end{equation}

The scales $s_{x, 16 \rightarrow 8}$ are computed offline using a calibration dataset.

\begin{figure}[htbp]
    \centering
    \includegraphics[width=1\linewidth]{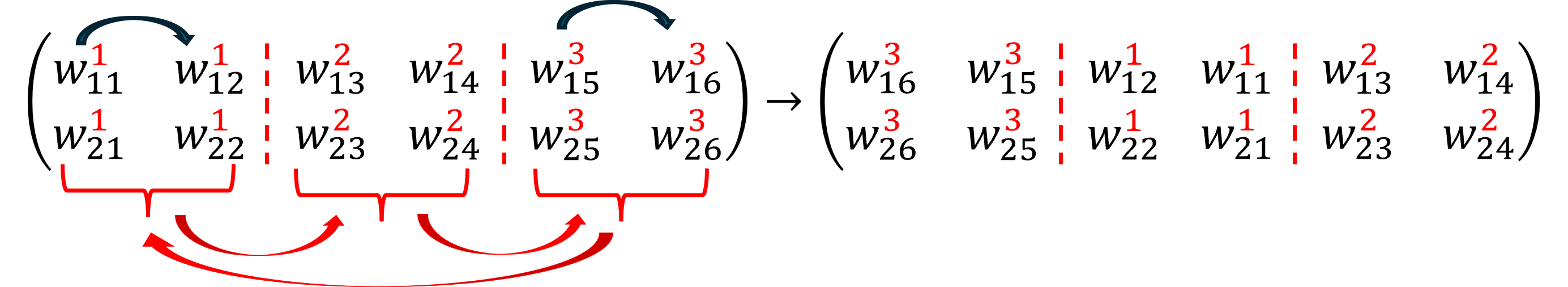}
    \caption{An illustration of the GAR method. In this example, the weight tensor is divided into three groups, indicated by the red superscript. The black arrows represent the local ordering within each group, based on the diagonal elements of the Hessian for that group, while the red arrows denote the global ordering between groups, determined by the maximum diagonal Hessian element of each group. At the end of the quantization process, the tensor is re-permuted to its original order. Notably, scales operate on consecutive weights after the ordering process (as opposed to activation reordering done in \cite{gptq}.)}
    \label{fig:GAR_illustration}
\end{figure}

\subsection{Inference-Aware Weight Reordering} \label{ssec:GAR}
A key challenge in the weight compensation approach is determining the optimal order for quantizing weights. The OBQ algorithm follows a greedy strategy, selecting the weight that introduces the least additional quantization error at each step. In contrast, GPTQ quantizes all rows in a predefined order. The original GPTQ paper proposes two schemes in the quantization process.
The first preserves the original order of weights within the tensor, and the second reorders weights based on the Hessian diagonal, prioritizing the quantization of "more important" weights.
While the second approach improves accuracy, it introduces inference overhead due to the need of memory indexing in weight grouping. Specifically, the weights $\textbf{W}_8^g$ are consecutive in the permuted tensor, and thus the scales $s_{8 \rightarrow 4}^g$ and zero-points $z_{8 \rightarrow4}^g$ operate on consecutive elements in the permuted tensor. However, the tensor saved and used in inference is the tensor in the original order, not the permuted tensor, and dequantizing online the weights using their pre-computed scales and zero-points can not be done in one multiplication as in the original case, thus increasing the inference latency (more details are given in \cref{sec:Sup-Gar}). 

To address this, we propose Group-Aware Reordering (GAR), a reordering scheme that leverages Hessian information to prioritize important weights during quantization while avoiding inference overhead. GAR reorders weights under the constraint that weights can be permuted in one of two ways - within a group, and/or by moving whole groups. With GAR, scales operate on consecutive weights in the original order. This ensures that group-wise scales and weight groups remain consistent throughout inference. 
Although GAR can also be beneficial for other methods, such as W4A16 GPTQ, it is especially helpful for DPQ. Without GAR, due to the nature of the predefined quantization order, alongside double quantization, large errors may accumulate in the weights which are quantized last. GAR reduces this error by quantizing the hard-to-quantize weights first, which then causes a reduced accumulated error to the latter weights.

\begin{figure}[t]
    \centering
    \includegraphics[width=1.1\linewidth]{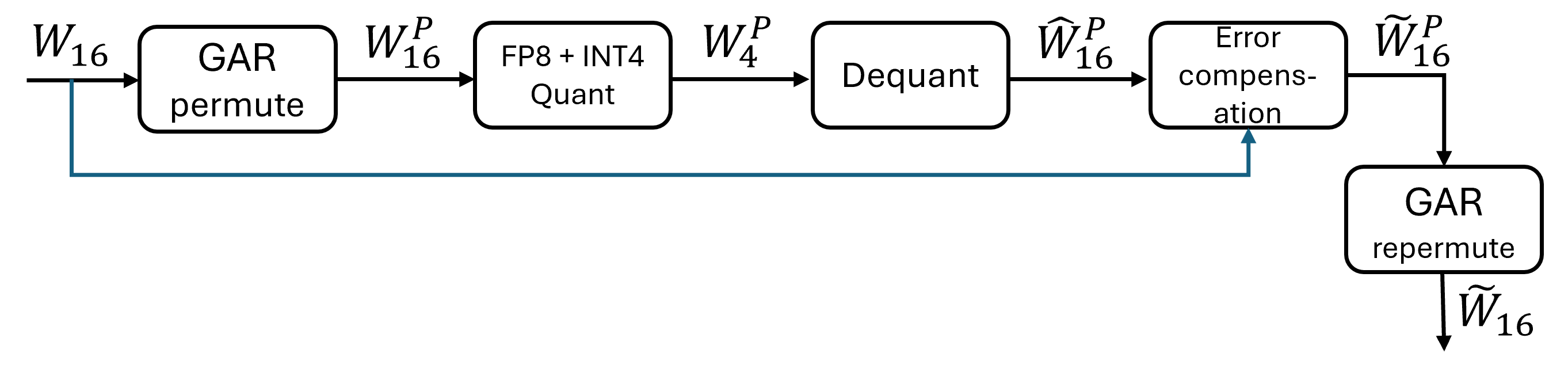}
    \caption{High-level illustration of the proposed DPQ algorithm. We begin by hybrid reordering (GAR) of the weights based on their importance. Next, we apply two quantization levels: FP8 and INT4. The quantization errors from both processes are computed and distributed to the next, less 'important' weights. Finally, we re-permute the tensor and the scales back to their original order.}
    \label{fig:DPAQ_illustration}
\end{figure}

\begin{figure}[t]
    \centering
    \includegraphics[width=0.9\linewidth]{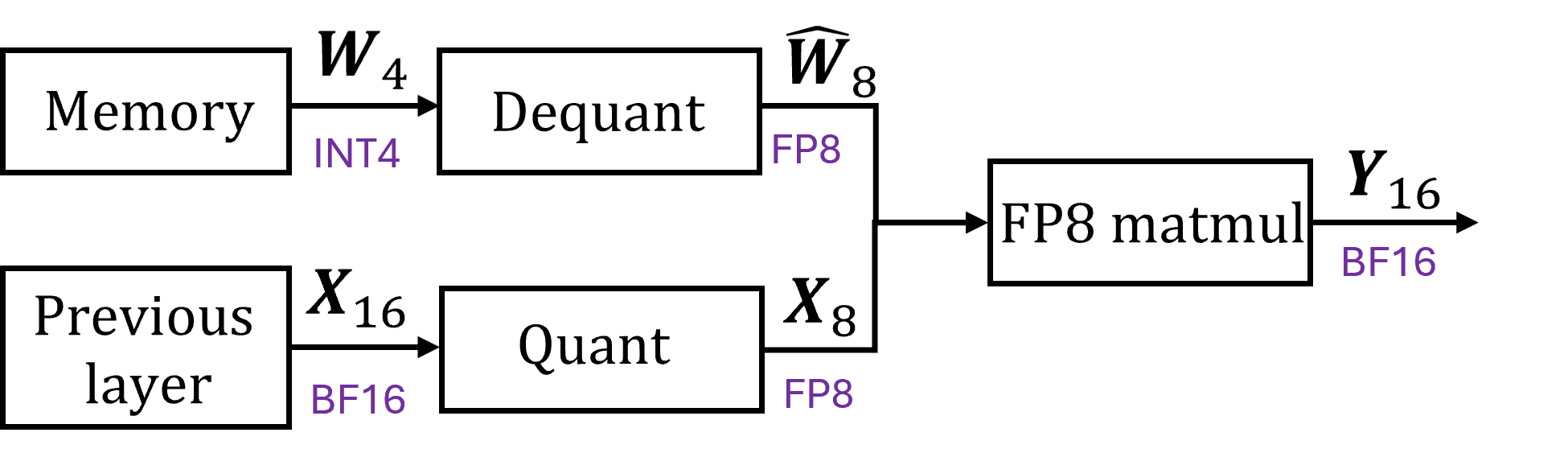}
    \caption{An illustration of the inference flow of our proposed W4A8 scheme.}
    \label{fig:model_illustration}
\end{figure}

In \cref{sec:Experiments} we demonstrate that our approach achieves higher accuracy compared to quantization without reordering and performs only slightly worse than unrestricted reordering while maintaining efficient inference execution.
An illustration of the GAR method is shown in \cref{fig:GAR_illustration}, and a detailed explanation of the scheme is provided in \cref{sec:Sup-Gar}.
\cref{fig:DPAQ_illustration} provides a high-level illustration of the proposed DPQ algorithm and  \cref{fig:model_illustration} provides an illustration of the inference flow under our W4A8 scheme.

\subsection{Further Optimizations and Considerations}
To further improve our quantization algorithm, we integrate additional techniques to enhance accuracy and efficiency.
First, the per-group scaling factor $s_{8 \rightarrow 4}^g$ is determined by searching for the scale that minimizes the mean squared error (MSE). Additionally, we apply clipping to constrain values within the dynamic range of the low-precision tensor.
As done in \cite{gptq}, we apply dampening to the Hessian matrix and employ the Cholesky decomposition for more efficient computation of the Hessian matrix in \cref{algorithm_update}.

\begin{algorithm2e}[h] 
\caption{Dual Precision Quantization (DPQ)}
\label{pseudo}
\SetAlgoLined
\KwIn{Weight matrix $\boldsymbol{W}_{16}$ in BF16, Hessian matrix $\boldsymbol{H}$}
\KwOut{Quantized weight matrix $\textbf{W}_4$, scaling factors $s_{16 \rightarrow 8}, s_{8 \rightarrow 4}^g$, zero-point $z_{ 8 \rightarrow 4}^g$}
    Compute scaling factor $s_{w, 16 \to 8}$ (\cref{scale_bf16_to_fp8}) \;
\ForEach{row $\boldsymbol{w} \in \boldsymbol{W}_{16}$}{    
    \ForEach{group $\mathcal{G}$ in $\boldsymbol{w}$}{
        Compute per-group scale $s_{8 \rightarrow 4}^g$ and zero-point $z_{8 \rightarrow 4}^g$ (\cref{scale_FP8_to_INT4}) \;

        \ForEach{weight $w_{16} \in \mathcal{G}$}{
            $w_{8} = \text{quant}(w_{16})$ (\cref{quantized w_8}) \;
            $w_{4} = \text{quant}(w_{8})$ (\cref{quantized w_4}) \;
            $\hat{w}_8 = \text{dequant}(w_4)$ (\cref{dequantized w_8}) \;
            $\hat{w}_{16} = \text{dequant}(\hat{w}_8)$ (\cref{dequantized w_16}) \;
            Compute the update $\delta_{F}$ (\cref{algorithm_update}) \;  
            Distribute $\delta_{F}$ to remaining weights (\cref{algorithm_update})
        }
    }
}
\end{algorithm2e}

\section{Experiments} \label{sec:Experiments}
In this section, we evaluate our proposed quantization and inference scheme in terms of both task accuracy and model performance (i.e., throughput). 
All experiments were conducted on Gaudi 2 and 3 devices, with BF16 precision used as the baseline for performance comparison. The quantization experiments were implemented using Intel’s Neural Compressor framework\footnote{https://github.com/intel/neural-compressor}, an open-source toolkit for neural network compression.
We begin by evaluating visual reasoning on the MMMU \cite{mmmu}, MMBench \cite{mmbench}, and Math Vista \cite{mathvista} benchmarks using the Qwen2-VL model family \cite{qwen2}, testing the 2B, 7B, and 72B models. 
The tasks were evaluated using VLMEvalkit \cite{duan2024vlmevalkit} on Intel Gaudi 2.
Next, since the language component is a crucial aspect of large vision-language models, and since previous W4A8 methods have focused on it, we also evaluate our method on the Llama-2 \cite{llama2} and Llama-3 \cite{llama3} model families.
We evaluate the accuracy of our language models across three distinct categories: WikiText-2 \cite{wikitext} for the measurement of perplexity, a comprehensive common sense reasoning suite, and MMLU \cite{mmlu} for the assessment of broad knowledge. The common sense reasoning evaluation is comprised of the following: HellaSwag (HS) \cite{hellaswag}, LAMBADA \cite{lambada}, BoolQ \cite{boolq}, ARC Easy (ARC-e) \cite{arceasy}, PIQA (PQ) \cite{piqa}, Winogrande (WG) \cite{winogrande}, ARC Challenge (Arc-c) \cite{arceasy}, and OpenBookQA \cite{openbookqa}. Results are reported as the average normalized accuracy for all tasks. The language tasks were evaluated using the LM-Evaluation-Harness framework \cite{eval-harness} in a zero-shot setting to ensure consistent comparison between different experiments. Our code is publicly available.

\subsection{Implementation Details}
The implementation of E4M3 in Gaudi 2 follows the IEEE standard for floating points, which reserves the largest exponent for NaN (not-a-number) and infinite values, and has the range of $\pm 240$. 
In Gaudi 3 the maximal exponent can be used for normal numbers, increasing the numerical range of E4M3 to $\pm 448$.

We used static (off-line) activation quantization, where the scaling factors are determined through calibration on the WebQuestions (WebQS) dataset \cite{webqs}, while the calibration set for the error compensation procedure in the weight quantization is a subset of \cite{gao2020pile800gbdatasetdiverse}.
In all our experiments, we use a group size of 128 for the INT4 asymmetric quantization, and a maximum sequence length of 2048. The FP8 scales are computed per tensor with symmetric quantization.

In Llama-2 and Llama-3 models, we quantize all linear layers to INT4, except for the embedding layer and the lm-head, as is common in other works. In the Qwen-VL model, we quantize all linear layer weights of the language model. Quantization times are similar to the timings reported in \cite{gptq}.

\subsection{Accuracy Results} \label{accuracy}
We use the WxAy notation to denote a quantization scheme in which weights are stored with $x$-bit precision and activations with $y$-bit precision. \Cref{tab:vision accuracy} presents the results for the multi-modal dataset evaluated on the Qwen2-VL model family.  The proposed DPQ algorithm demonstrates consistent performance across different model sizes and tasks, with only a minor degradation compared to the full BF16 reference. Notably, for most tasks, it outperforms the W4A16 GPTQ algorithm despite operating at a lower precision. This improvement can be attributed to the GAR method, which enhances model accuracy.

We next present a comparison of the proposed DPQ algorithm with state-of-the-art methods, including QServe \cite{qserve}, QQQ \cite{qqq}, GPTQ \cite{gptq}, and the vanilla round-to-nearest (RTN) method on language tasks. The results for QServe and QQQ are taken directly from their papers, where we used the 128-group size configuration as in DPQ for a fair comparison. The GPTQ method is implemented without the FP8 error compensation used in DPQ (i.e., only the INT4 error is compensated).
\Cref{tab:comparison} demonstrates that DPQ outperforms both RTN and the naive implementation of GPTQ, as expected. Additionally, DPQ achieves better results than QServe and QQQ, which follow a different approach (both perform quantization to INT8 rather than FP8).
We also note that many optimizations in the QServe algorithm, such as block input/output module rotation and smoothing, are orthogonal to our method and could be integrated into DPQ to further increase its accuracy.

\Cref{tab:accuracy} presents the results for the WikiText-2, reasoning, and common sense tasks under various precisions. 
As expected, schemes that quantize weights to 4 bits exhibit greater accuracy degradation compared to the W8A8 algorithm, due to the significantly lower bit representation. However, our proposed W4A8 DPQ algorithm achieves a reasonable accuracy degradation compared to the full precision model, and only a small accuracy degradation compared to the W4A16 scheme (in \cref{tab:accuracy} we implemented GPTQ with GAR, thus it serves as a natural upper bound for DPQ) while maintaining a more efficient trade-off between accuracy and latency (detailed in \cref{ssec:performance}).

\begin{table*}[htbp]
\centering
\scalebox{0.7}{
\resizebox{\textwidth}{!}{%
    \begin{tabular}{|l|l|cc|cc|cc|cc|} 
        \hline
        Model & Configuration & \multicolumn{2}{c|}{MMMU} & \multicolumn{2}{c|}{MMBench} & \multicolumn{2}{c|}{MathVista} \\
        & & Acc~$\uparrow$ & $\Delta$ (\%)~$\uparrow$ & Acc~$\uparrow$ & $\Delta$ (\%)~$\uparrow$ & Acc~$\uparrow$ & $\Delta$ (\%)~$\uparrow$ \\
        \hline
        \multirow{3}{*}{Qwen2-VL-2B-Instruct} & BF16 Reference & 41.88 & --  & 72.07 & -- & 44.40 & -- \\
        & W4A16 (GPTQ) & 39.22 & -6.78  & 70.87 & -1.69 & 41.69 & -6.50 \\
        & W4A8 (DPQ) & 39.37 & -6.37  & 71.12 & -1.33 & 45.3 & +1.98 \\
        \hline
        \multirow{3}{*}{Qwen2-VL-7B-Instruct} & BF16 Reference & 	53.77 & -- & 81.78 & -- & 58.20 & -- \\
        & W4A16 (GPTQ) &52.55 & -2.32 & 81.27 & -0.63 & 60.30 & +3.48 \\
        & W4A8 (DPQ) & 52.64 & -2.14  & 81.25 & -0.65 & 59.86 & -2.77 \\
        \hline
        \multirow{3}{*}{Qwen2-VL-72B-Instruct} & BF16 Reference & 	65.44 & -- & 86.94 & -- & 70.19 & --  \\
        & W4A16 (GPTQ) & 64.00 & -2.25 & 86.68 & -0.30 & 69.20 & -1.43 \\
        & W4A8 (DPQ)  & 63.98 & -2.28 & 86.48 & -0.53 & 68.97 & -1.77 \\
        \hline
    \end{tabular}%
    }
    }
     \caption{Accuracy of vision tasks under various precisions. BF16 and GPTQ results are taken from the Huggingface model pages.}
     \label{tab:vision accuracy}
\end{table*}

\begin{table*}[htbp]
\centering
\scalebox{0.8}{
\resizebox{\textwidth}{!}{%
    \begin{tabular}{|l|cc|cc|cc|cc|cc|cc|cc|cc|cc|cc} 
        \hline
        Algorithm & \multicolumn{2}{c|}{AVG} & \multicolumn{2}{c|}{ARC-e} & \multicolumn{2}{c|}{ARC-c} & \multicolumn{2}{c|}{HS} & \multicolumn{2}{c|}{WG} & \multicolumn{2}{c|}{PQ} \\
         & Acc~$\uparrow$ & $\Delta$ (\%)~$\uparrow$ & Acc~$\uparrow$ & $\Delta$ (\%)~$\uparrow$ & Acc~$\uparrow$ & $\Delta$ (\%)~$\uparrow$ & Acc~$\uparrow$ & $\Delta$ (\%)~$\uparrow$  & Acc~$\uparrow$ & $\Delta$ (\%)~$\uparrow$ & Acc~$\uparrow$ & $\Delta$ (\%)~$\uparrow$\\
        \hline
        BF16 & 68.64  & -- & 73.86  & --  & 44.88  & -- & 76.13 & -- & 69.29 & -- & 79.05 & -- \\
        DPQ (ours) & \textbf{68.15} & \textbf{-0.73} & 72.81  & -1.45 & \textbf{45.13} & \textbf{+0.55} &\textbf{75.17}& \textbf{-1.28} &\textbf{69.45}&  \textbf{+0.22} & 78.18 & -1.11 \\
        QServe & 67.95 & -1.02 & \textbf{73.32} & \textbf{-0.74} & 44.80  & -0.18 &74.98& -1.53&68.59&-1.03&78.07&-1.26 \\
        QQQ & 67.47 & -1.74 & 72.94 & -1.27 &44.37 & -1.15  &74.53&-2.15&67.01 &-3.41&\textbf{78.51}& \textbf{-0.69}\\
        GPTQ & 64.69 &-6.11  &67.88  & -8.82 & 42.15  &-6.48  &71.64&-6.27&65.27&-6.17&76.55&-3.27\\
        RTN & 	63.54 & -8.03 &66.87  &-10.46  &40.87  & -9.87 &71.31&-6.76&63.53&-9.08&75.13&-5.22\\     
        \hline
    \end{tabular}%
}
}
     \caption{Accuracy under various quantization algorithms for W4A8 scheme on Llama2-7B. The GPTQ results are an adaptation of GPTQ to W4A8, without accounting the FP8 quantization error. The results for QServe and QQQ were taken from \cite{qserve} and \cite{qqq}, respectively.}
     \label{tab:comparison}
\end{table*}

\begin{table*}[htbp]
\centering
\scalebox{0.65}{
\resizebox{\textwidth}{!}{%
    \begin{tabular}{|l|l|cc|cc|cc|} 
        \hline
        Model & Configuration & \multicolumn{2}{c|}{WikiText2} & \multicolumn{2}{c|}{Commonsense} & \multicolumn{2}{c}{MMLU} \\
        & & Ppl~$\downarrow$ & $\Delta$ (\%)~$\downarrow$ & Acc~$\uparrow$ & $\Delta$ (\%)~$\uparrow$ & Acc~$\uparrow$ & $\Delta$ (\%)~$\uparrow$ \\
        \hline
        \multirow{5}{*}{Llama-2-7B} & BF16 Reference & 5.472 & -- & 65.851 & -- & 43.074 & -- \\
        & W8A8 & 5.502 & +0.55 & 65.679 & -0.26 & 43.006 & -0.16 \\
        & W4A16* (GPTQ) & 5.629 & +2.79 & 65.049 & -1.22 & 42.891 & -0.48 \\
        & W4A8* (DPQ) & 	5.662 & +3.36 & 65.582	 & -0.41 & 42.628 & -1.04 \\
        & W4A16 (GPTQ) & 5.745 & +4.75 & 	65.324 & -0.81 & 41.662 & -3.28 \\
        & W4A8 (DPQ) & 5.781 & +5.34 & 64.269 & -2.40 & 41.931 & -2.65 \\
        \hline
        \multirow{5}{*}{Llama-3.1-8B} & BF16 Reference & 	6.238 & -- &70.809	 & -- & 66.626 & -- \\
        & W8A8 & 6.311	& +1.15 &  70.202	& -0.86	& 66.021	& -0.91    \\
        & W4A16* (GPTQ) &6.597  & +5.45 &	69.657 &	-1.63 &	65.223	& -2.11  \\
        & W4A8*(DPQ) & 6.935 & +10.04 & 69.588	&-1.72&	65.206&	-2.13  \\
        & W4A16 (GPTQ) & 6.631 & +5.93 & 69.635&	-2.04&	64.073&	-3.83 \\
        & W4A8 (DPQ) & 6.704& +6.95 &69.074 &-2.45& 63.987& -3.96 \\
        \hline
        \multirow{3}{*}{Llama-2-70B} & BF16 Reference & 3.320
        & -- &73.561 & -- & 67.691 & -- \\
        & W4A16 (GPTQ) & 3.420 & +2.92 & 73.159 & -0.55 & 66.828 & -1.28 \\
        & W4A8 (DPQ) & 3.466 & +4.21 & 73.129 & -0.59 & 66.346 & -1.99 \\
        \hline
    \end{tabular}%
    }}
     \caption{Accuracy of language tasks under various precisions. The notations $W4A16^*$ and $W4A8^*$ refer to schemes that use full activation reordering, which reduces accuracy degradation at the cost of increased latency. GPTQ (W4A16) is included in the results, since it is a common quantization algorithm. Also, since DPQ adds an additional level of quantization over GPTQ, it serves as an upper-bound to DPQ.}
     \label{tab:accuracy}
\end{table*}

\subsection{Performance Analysis} \label{ssec:performance}
Using W4A8 improves model performance in several ways. 
With respect to full FP8 quantization (weights and activations - W8A8) - using 4 bit weights allows to further reduce the memory footprint, thus allowing for larger models to fit into smaller or less devices (when using model parallel), and improves the memory bandwidth.
For example, using 4-bit weights allows models such as Llama 3.2 90B Vision to fit into a device such as Intel Gaudi 2 (96 Gb HBM) or Nvidia H100 (80Gb HBM).
With respect to 4-bit weight-only quantization (W4A16), using FP8 precision for the activations enables larger batch sizes (which yields faster overall tokens per second), and for faster GeMM operations, if the hardware has support for it (Gaudi 2, Gaudi 3, Nvidia H100, H200 have support for such).

\begin{figure}[t]
    \centering
    \includegraphics[width=1.1\linewidth]{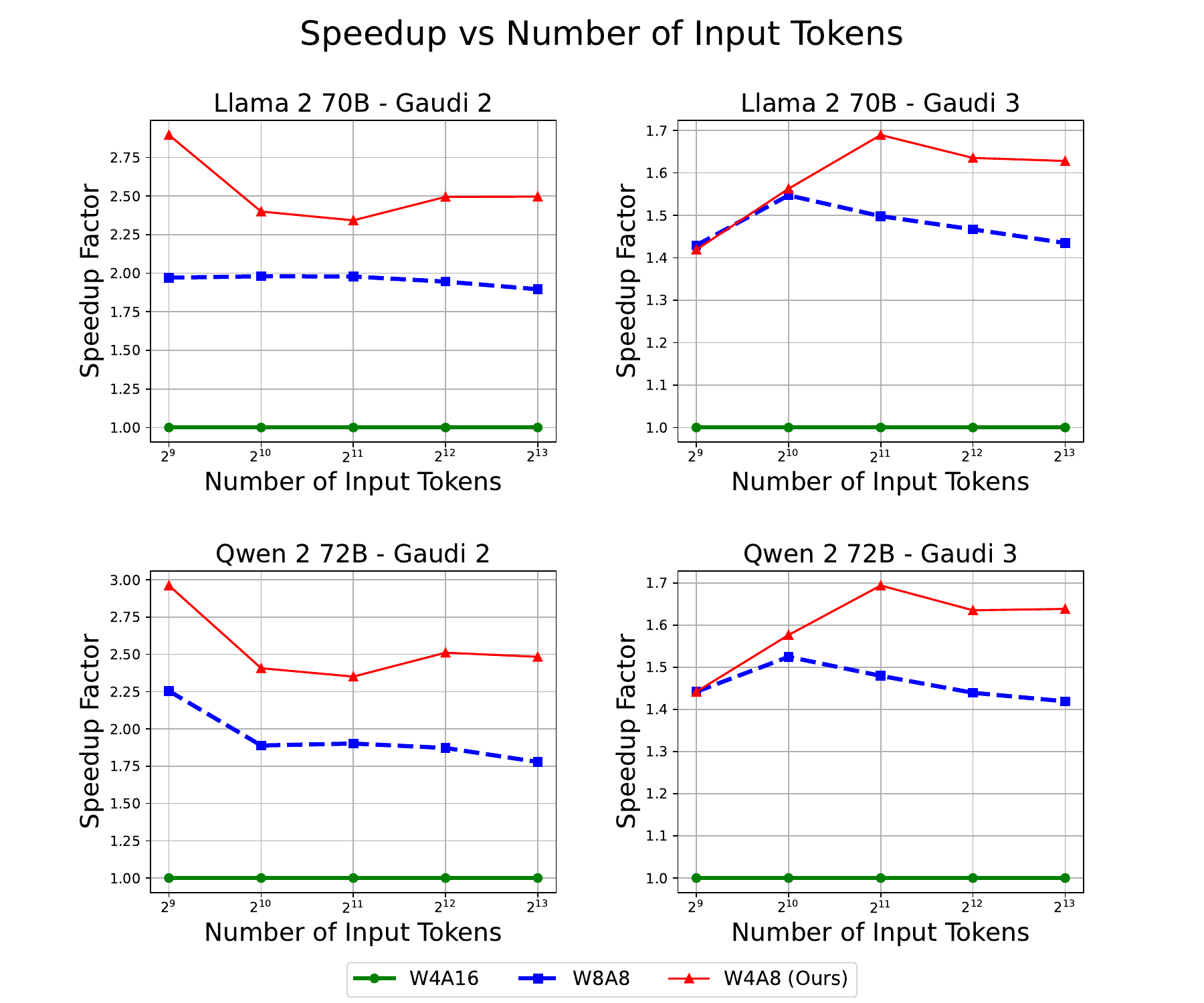}
    \caption{Speed-up comparison between 3 methods on Gaudi 2 and Gaudi 3 accelerators. Each subplot compares speed-up of W4A8 (red) and W8A8 (blue) compared to W4A16 (green). The two upper plots compare Llama 2 70B on Gaudi 2 (left) and Gaudi 3 (right). The bottom plots compare Qwen 2 72B on Gaudi 2 (left) and Gaudi 3 (right). As can be seen, W4A8 can reach up to 3x speed-up over W4A16, and up to 1.4x speed-up over W8A8. W4A8 uses GAR, thus weights are consecutive in memory during inference. }
    \label{fig:fig_speedup}
    \vspace{-5pt}
\end{figure}

To showcase the potential gains for the W4A8 format, we provide LLM model projections in \cref{fig:fig_speedup}. We analyze performance for various input and output lengths. Since many recent VLMs use both a Vision Transformer and an LLM, we focus on LLM performance, since the majority of compute and runtime is due to the auto-regressive LLM, rather than the ViT which runs once and is a fraction of the model.
We compare both 4 bit weights (W4A16) and FP8 quantization (W8A8), since both are popular quantization choices. Note that for the chosen models, a W16A16 (native BF16) model does not fit on a single device, thus not shown.
The speed-up is achieved by larger batch sizes (which are enabled by smaller model weights), as well as more efficient compute, due to the FP8 GeMM engines being used.

\subsection{Ablation Study}
We conduct an ablation study to evaluate the impact of our proposed activation hybrid reordering GAR method on model accuracy. As discussed earlier, we expect this hybrid scheme to outperform the scheme without reordering. Although our method imposes constraints on the activation order, we note that most layers are characterized by a small number of highly important output activations \cite{super, massive}. Consequently, even though our scheme only partially orders activations, it effectively captures the most significant weights. Thus, we expect only a slight accuracy degradation compared to the full reordering scheme.

Our results, presented in \Cref{tab:ablation}, support this hypothesis. Under both W4A16 and W4A8 quantization schemes, GAR yields a notable improvement over the scheme without reordering, while exhibiting only a minor accuracy degradation compared to full reordering scheme. 
Note that GAR requires a ranking criterion to determine which group should be quantized first. In our experiments, we ordered groups based on their maximum Hessian diagonal element. Alternative criteria are possible, for example, ranking groups by the average of the top 10\% largest elements before sorting. We leave the exploration of other ranking strategies for future work.

Another key insight into our algorithm emerges from the improved accuracy of DPQ compared to the naive implementation of GPTQ, as shown in \cref{tab:comparison}. DPQ mitigates nearly $4\%$ of the accuracy degradation relative to the full-precision reference. This is because our method compensates for the error using the dequantized weight from both quantization levels, allowing it to account for both sources of error and produce a more accurate layer output. In contrast, a naive implementation of GPTQ in the W4A8 scheme introduces the FP8 quantization error on top of the compensated INT4 quantization error, resulting in the degradation observed in \cref{tab:comparison}.

\begin{table}[htbp]
\centering
\scalebox{1.02}{
\resizebox{\columnwidth}{!}{%
    \begin{tabular}{l|l|cc|cc|cc} 
        \hline
        Prec. & Config. & \multicolumn{2}{c|}{PPL - WikiText2} & \multicolumn{2}{c|}{Commonsense} & \multicolumn{2}{c}{MMLU} \\
        & & Acc~$\downarrow$ & $\Delta$ (\%)~$\downarrow$ & Acc~$\uparrow$ & $\Delta$ (\%)~$\uparrow$ & Acc~$\uparrow$ & $\Delta$ (\%)~$\uparrow$ \\
        \hline
        \multirow{5}{*}{ W4A16} & BF16 & 6.238 & -- &70.809	 & -- & 66.626 & --  \\
        & Ordered & 6.597 & +5.45 & 69.657 & -1.63 & 65.223 & -2.11 \\
        & GAR (ours) & 6.600 & +5.48 & 70.319 & -0.69 & 65.114 & -2.27 \\
        & Unordered & 	7.680 & +18.78 &69.285 & -2.15 & 63.499 & -4.69 \\
        \hline
        \multirow{5}{*}{  W4A8} & BF16 & 6.238 & -- &70.809	 & -- & 66.626 & -- \\
        & Ordered & 6.679 & +6.59 &	69.239 & -2.22 & 64.368 & -3.39 \\
        & GAR (ours) & 6.704 & +6.95 &	69.074 & -2.45 & 63.987 & -3.96 \\
        & Unordered & 32.812 & +80.99 & 69.026 & -2.52 & 61.716 &	-7.37 \\
        \hline
    \end{tabular}%
    }
    }
     \caption{Accuracy under different activation ordering methods for Llama-3.1-8B. 'Ordered’ refers to full activation reordering, ‘Unordered’ means preserving the tensor in its original form.
     The scales for all methods are selected to minimize the mean squared quantization error between the dequantized and original tensors.} 
     \label{tab:ablation}
\end{table}

\subsection{Discussion} \label{discussion}
We next discuss some key points regarding our work.

\begin{enumerate}
    \item The W4A16 scheme (when implemented with GAR as in \cref{tab:accuracy}), which only quantizes weights and performs computation in BF16, serves as a natural upper bound for our W4A8 approach . That is, we primarily compare our accuracy degradation against W4A16. 
    Optimizing W4A16 (weight-only) is beyond the scope of this paper.

    \item  Our quantization method is orthogonal to many existing quantization strategies, such as AWQ, QuaRot, and others. Incorporating these techniques may further enhance our results. We leave this for future work.

\end{enumerate}

\section{Conclusion} \label{sec:conclusion}
In this paper, we introduce DPQ, a dual precision aware W4A8 quantization algorithm designed for efficient inference on modern accelerators. Our method leverages a two-step quantization process, mapping high-precision weights to INT4 and activations to FP8, enabling efficient FP8 matrix multiplication while minimizing accuracy degradation. We further enhance our approach with error compensation technique and Hessian-guided weight reordering, ensuring a good balance between performance and accuracy.
Experimental analysis on Qwen2-VL, Llama-2, and Llama-3 demonstrate that our W4A8 quantization method achieves significant throughput improvements while maintaining competitive accuracy relative to reference schemes. 

{
    \small
    \bibliographystyle{IEEEtran}
    \bibliography{main}

\begin{thebibliography}{10}
\providecommand{\url}[1]{#1}
\csname url@samestyle\endcsname
\providecommand{\newblock}{\relax}
\providecommand{\bibinfo}[2]{#2}
\providecommand{\BIBentrySTDinterwordspacing}{\spaceskip=0pt\relax}
\providecommand{\BIBentryALTinterwordstretchfactor}{4}
\providecommand{\BIBentryALTinterwordspacing}{\spaceskip=\fontdimen2\font plus
\BIBentryALTinterwordstretchfactor\fontdimen3\font minus \fontdimen4\font\relax}
\providecommand{\BIBforeignlanguage}[2]{{%
\expandafter\ifx\csname l@#1\endcsname\relax
\typeout{** WARNING: IEEEtran.bst: No hyphenation pattern has been}%
\typeout{** loaded for the language `#1'. Using the pattern for}%
\typeout{** the default language instead.}%
\else
\language=\csname l@#1\endcsname
\fi
#2}}
\providecommand{\BIBdecl}{\relax}
\BIBdecl

\bibitem{QAT1}
S.~K. Esser, J.~L. McKinstry, D.~Bablani, R.~Appuswamy, and D.~S. Modha, ``Learned step size quantization,'' \emph{arXiv preprint arXiv:1902.08153}, 2019.

\bibitem{QAT2}
B.~Zhuang, M.~Tan, J.~Liu, L.~Liu, I.~Reid, and C.~Shen, ``Effective training of convolutional neural networks with low-bitwidth weights and activations,'' \emph{IEEE Transactions on Pattern Analysis and Machine Intelligence}, vol.~44, no.~10, pp. 6140--6152, 2021.

\bibitem{QAT3}
J.~H. Lee, J.~Yun, S.~J. Hwang, and E.~Yang, ``Cluster-promoting quantization with bit-drop for minimizing network quantization loss,'' in \emph{Proceedings of the IEEE/CVF International Conference on Computer Vision}, 2021, pp. 5370--5379.

\bibitem{QAT4}
Z.~Liu, B.~Oguz, C.~Zhao, E.~Chang, P.~Stock, Y.~Mehdad, Y.~Shi, R.~Krishnamoorthi, and V.~Chandra, ``Llm-qat: Data-free quantization aware training for large language models,'' \emph{arXiv preprint arXiv:2305.17888}, 2023.

\bibitem{nagel2019data}
M.~Nagel, M.~v. Baalen, T.~Blankevoort, and M.~Welling, ``Data-free quantization through weight equalization and bias correction,'' in \emph{Proceedings of the IEEE/CVF international conference on computer vision}, 2019, pp. 1325--1334.

\bibitem{hubara2020improving}
I.~Hubara, Y.~Nahshan, Y.~Hanani, R.~Banner, and D.~Soudry, ``Improving post training neural quantization: Layer-wise calibration and integer programming,'' \emph{arXiv preprint arXiv:2006.10518}, 2020.

\bibitem{hubara2021accurate}
I.~Hubara, Y.~Nahshan, Y.~Hanani, R.~Banner, and D.Soudry, ``Accurate post training quantization with small calibration sets,'' in \emph{International conference on machine learning}.\hskip 1em plus 0.5em minus 0.4em\relax PMLR, 2021, pp. 4466--4475.

\bibitem{liu2021post}
Z.~Liu, Y.~Wang, K.~Han, W.~Zhang, S.~Ma, and W.~Gao, ``Post-training quantization for vision transformer,'' \emph{Advances in Neural Information Processing Systems}, vol.~34, pp. 28\,092--28\,103, 2021.

\bibitem{nahshan2021loss}
Y.~Nahshan, B.~Chmiel, C.~Baskin, E.~Zheltonozhskii, R.~Banner, A.~M. Bronstein, and A.~Mendelson, ``Loss aware post-training quantization,'' \emph{Machine Learning}, vol. 110, no.~11, pp. 3245--3262, 2021.

\bibitem{ZeroQuant}
Z.~Yao, R.~Yazdani~Aminabadi, M.~Zhang, X.~Wu, C.~Li, and Y.~He, ``Zeroquant: Efficient and affordable post-training quantization for large-scale transformers,'' \emph{Advances in Neural Information Processing Systems}, vol.~35, pp. 27\,168--27\,183, 2022.

\bibitem{AdaRound}
M.~Nagel, R.~A. Amjad, M.~Van~Baalen, C.~Louizos, and T.~Blankevoort, ``Up or down? adaptive rounding for post-training quantization,'' in \emph{International conference on machine learning}.\hskip 1em plus 0.5em minus 0.4em\relax PMLR, 2020, pp. 7197--7206.

\bibitem{gptq}
E.~Frantar, S.~Ashkboos, T.~Hoefler, and D.~Alistarh, ``Gptq: Accurate post-training quantization for generative pre-trained transformers,'' \emph{arXiv preprint arXiv:2210.17323}, 2022.

\bibitem{awq}
J.~Lin, J.~Tang, H.~Tang, S.~Yang, W.-M. Chen, W.-C. Wang, G.~Xiao, X.~Dang, C.~Gan, and S.~Han, ``Awq: Activation-aware weight quantization for on-device llm compression and acceleration,'' \emph{Proceedings of Machine Learning and Systems}, vol.~6, pp. 87--100, 2024.

\bibitem{VISION1}
C.~Wang, Z.~Wang, X.~Xu, Y.~Tang, J.~Zhou, and J.~Lu, ``Q-vlm: Post-training quantization for large vision-language models,'' \emph{Advances in Neural Information Processing Systems}, vol.~37, pp. 114\,553--114\,573, 2025.

\bibitem{VISION2}
J.~Xie, Y.~Zhang, M.~Lin, L.~Cao, and R.~Ji, ``Advancing multimodal large language models with quantization-aware scale learning for efficient adaptation,'' in \emph{Proceedings of the 32nd ACM International Conference on Multimedia}, 2024, pp. 10\,582--10\,591.

\bibitem{CVPR1}
Y.~Ding, W.~Feng, C.~Chen, J.~Guo, and X.~Liu, ``Reg-ptq: Regression-specialized post-training quantization for fully quantized object detector,'' in \emph{Proceedings of the IEEE/CVF Conference on Computer Vision and Pattern Recognition}, 2024, pp. 16\,174--16\,184.

\bibitem{CVPR4}
C.~Lv, H.~Chen, J.~Guo, Y.~Ding, and X.~Liu, ``Ptq4sam: Post-training quantization for segment anything,'' in \emph{Proceedings of the IEEE/CVF Conference on Computer Vision and Pattern Recognition}, 2024, pp. 15\,941--15\,951.

\bibitem{fp8formates}
P.~Micikevicius, D.~Stosic, N.~Burgess, M.~Cornea, P.~Dubey, R.~Grisenthwaite, S.~Ha, A.~Heinecke, P.~Judd, J.~Kamalu \emph{et~al.}, ``Fp8 formats for deep learning,'' \emph{arXiv preprint arXiv:2209.05433}, 2022.

\bibitem{lee2025faster}
J.~Lee, S.~Markovich-Golan, D.~Ohayon, Y.~Hanani, G.~Park, B.~Kim, A.~Karnieli, U.~Livne, H.~Shen, T.~Huang \emph{et~al.}, ``Faster inference of llms using fp8 on the intel gaudi,'' \emph{arXiv preprint arXiv:2503.09975}, 2025.

\bibitem{fp8_exponent}
A.~Kuzmin, M.~Van~Baalen, Y.~Ren, M.~Nagel, J.~Peters, and T.~Blankevoort, ``Fp8 quantization: The power of the exponent,'' \emph{Advances in Neural Information Processing Systems}, vol.~35, pp. 14\,651--14\,662, 2022.

\bibitem{kim2025investigation}
J.~Kim, J.~Lee, G.~Park, B.~Kim, S.~J. Kwon, D.~Lee, and Y.~Lee, ``An investigation of fp8 across accelerators for llm inference,'' \emph{arXiv preprint arXiv:2502.01070}, 2025.

\bibitem{FP8_versus_INT8}
M.~Van~Baalen, A.~Kuzmin, S.~S. Nair, Y.~Ren, E.~Mahurin, C.~Patel, S.~Subramanian, S.~Lee, M.~Nagel, J.~Soriaga \emph{et~al.}, ``Fp8 versus int8 for efficient deep learning inference,'' \emph{arXiv preprint arXiv:2303.17951}, 2023.

\bibitem{qserve}
Y.~Lin, H.~Tang, S.~Yang, Z.~Zhang, G.~Xiao, C.~Gan, and S.~Han, ``Qserve: W4a8kv4 quantization and system co-design for efficient llm serving,'' \emph{arXiv preprint arXiv:2405.04532}, 2024.

\bibitem{super}
M.~Yu, D.~Wang, Q.~Shan, and A.~Wan, ``The super weight in large language models,'' \emph{arXiv preprint arXiv:2411.07191}, 2024.

\bibitem{smoothquant}
G.~Xiao, J.~Lin, M.~Seznec, H.~Wu, J.~Demouth, and S.~Han, ``Smoothquant: Accurate and efficient post-training quantization for large language models,'' in \emph{International Conference on Machine Learning}.\hskip 1em plus 0.5em minus 0.4em\relax PMLR, 2023, pp. 38\,087--38\,099.

\bibitem{quarot}
S.~Ashkboos, A.~Mohtashami, M.~L. Croci, B.~Li, M.~Jaggi, D.~Alistarh, T.~Hoefler, and J.~Hensman, ``Quarot: Outlier-free 4-bit inference in rotated llms,'' \emph{arXiv preprint arXiv:2404.00456}, 2024.

\bibitem{squeezellm}
S.~Kim, C.~Hooper, A.~Gholami, Z.~Dong, X.~Li, S.~Shen, M.~W. Mahoney, and K.~Keutzer, ``Squeezellm: Dense-and-sparse quantization,'' \emph{arXiv preprint arXiv:2306.07629}, 2023.

\bibitem{ding2022towards}
Y.~Ding, H.~Qin, Q.~Yan, Z.~Chai, J.~Liu, X.~Wei, and X.~Liu, ``Towards accurate post-training quantization for vision transformer,'' in \emph{Proceedings of the 30th ACM international conference on multimedia}, 2022, pp. 5380--5388.

\bibitem{Noisyquant}
Y.~Liu, H.~Yang, Z.~Dong, K.~Keutzer, L.~Du, and S.~Zhang, ``Noisyquant: Noisy bias-enhanced post-training activation quantization for vision transformers,'' in \emph{Proceedings of the IEEE/CVF Conference on Computer Vision and Pattern Recognition}, 2023, pp. 20\,321--20\,330.

\bibitem{AutoRound}
W.~Cheng, W.~Zhang, H.~Shen, Y.~Cai, X.~He, K.~Lv, and Y.~Liu, ``Optimize weight rounding via signed gradient descent for the quantization of llms,'' \emph{arXiv preprint arXiv:2309.05516}, 2023.

\bibitem{OmniQuant}
W.~Shao, M.~Chen, Z.~Zhang, P.~Xu, L.~Zhao, Z.~Li, K.~Zhang, P.~Gao, Y.~Qiao, and P.~Luo, ``Omniquant: Omnidirectionally calibrated quantization for large language models,'' \emph{arXiv preprint arXiv:2308.13137}, 2023.

\bibitem{HQQ}
H.~Badri and A.~Shaji, ``Half-quadratic quantization of large machine learning models,'' \emph{Dan Hendrycks, Collin Burns, Steven Basart, Andy Zou, Mantas Mazeika, Dawn Song, and Jacob}, 2023.

\bibitem{PTQ_CVPR}
J.~Liu, L.~Niu, Z.~Yuan, D.~Yang, X.~Wang, and L.Wenyu, ``Pd-quant: Post-training quantization based on prediction difference metric,'' in \emph{Proceedings of the IEEE/CVF Conference on Computer Vision and Pattern Recognition}, 2023, pp. 24\,427--24\,437.

\bibitem{Pdquant}
J.~Liu, L.~Niu, Z.~Yuan, D.~Yang, X.~Wang, and W.~Liu, ``Pd-quant: Post-training quantization based on prediction difference metric,'' in \emph{Proceedings of the IEEE/CVF Conference on Computer Vision and Pattern Recognition}, 2023, pp. 24\,427--24\,437.

\bibitem{optimalbrainsurgery}
B.~Hassibi, D.~G. Stork, and G.~J. Wolff, ``Optimal brain surgeon and general network pruning,'' in \emph{IEEE international conference on neural networks}.\hskip 1em plus 0.5em minus 0.4em\relax IEEE, 1993, pp. 293--299.

\bibitem{OBQ}
E.~Frantar and D.~Alistarh, ``Optimal brain compression: A framework for accurate post-training quantization and pruning,'' \emph{Advances in Neural Information Processing Systems}, vol.~35, pp. 4475--4488, 2022.

\bibitem{yuan2022ptq4vit}
Z.~Yuan, C.~Xue, Y.~Chen, Q.~Wu, and G.~Sun, ``Ptq4vit: Post-training quantization for vision transformers with twin uniform quantization,'' in \emph{European conference on computer vision}.\hskip 1em plus 0.5em minus 0.4em\relax Springer, 2022, pp. 191--207.

\bibitem{qqq}
Y.~Zhang, P.~Zhang, M.~Huang, J.~Xiang, Y.~Wang, C.~Wang, Y.~Zhang, L.~Yu, C.~Liu, and W.~Lin, ``Qqq: Quality quattuor-bit quantization for large language models,'' \emph{arXiv preprint arXiv:2406.09904}, 2024.

\bibitem{qlora}
T.~Dettmers, A.~Pagnoni, A.~Holtzman, and L.~Zettlemoyer, ``Qlora: Efficient finetuning of quantized llms,'' \emph{Advances in neural information processing systems}, vol.~36, pp. 10\,088--10\,115, 2023.

\bibitem{SpinQuant}
Z.~Liu, C.~Zhao, I.~Fedorov, B.~Soran, D.~Choudhary, R.~Krishnamoorthi, V.~Chandra, Y.~Tian, and T.~Blankevoort, ``Spinquant: Llm quantization with learned rotations,'' \emph{arXiv preprint arXiv:2405.16406}, 2024.

\bibitem{massive}
M.~Sun, X.~Chen, J.~Z. Kolter, and Z.~Liu, ``Massive activations in large language models,'' \emph{arXiv preprint arXiv:2402.17762}, 2024.

\bibitem{mmmu}
X.~Yue, Y.~Ni, K.~Zhang, T.~Zheng, R.~Liu, G.~Zhang, S.~Stevens, D.~Jiang, W.~Ren, Y.~Sun \emph{et~al.}, ``Mmmu: A massive multi-discipline multimodal understanding and reasoning benchmark for expert agi,'' in \emph{Proceedings of the IEEE/CVF Conference on Computer Vision and Pattern Recognition}, 2024, pp. 9556--9567.

\bibitem{mmbench}
Y.~Liu, H.~Duan, Y.~Zhang, B.~Li, S.~Zhang, W.~Zhao, Y.~Yuan, J.~Wang, C.~He, Z.~Liu \emph{et~al.}, ``Mmbench: Is your multi-modal model an all-around player?'' in \emph{European conference on computer vision}.\hskip 1em plus 0.5em minus 0.4em\relax Springer, 2024, pp. 216--233.

\bibitem{mathvista}
P.~Lu, H.~Bansal, T.~Xia, J.~Liu, C.~Li, H.~Hajishirzi, H.~Cheng, K.-W. Chang, M.~Galley, and J.~Gao, ``Mathvista: Evaluating mathematical reasoning of foundation models in visual contexts,'' \emph{arXiv preprint arXiv:2310.02255}, 2023.

\bibitem{qwen2}
P.~Wang, S.~Bai, S.~Tan, S.~Wang, Z.~Fan, J.~Bai, K.~Chen, X.~Liu, J.~Wang, W.~Ge \emph{et~al.}, ``Qwen2-vl: Enhancing vision-language model's perception of the world at any resolution,'' \emph{arXiv preprint arXiv:2409.12191}, 2024.

\bibitem{duan2024vlmevalkit}
H.~Duan, J.~Yang, Y.~Qiao, X.~Fang, L.~Chen, Y.~Liu, X.~Dong, Y.~Zang, P.~Zhang, J.~Wang \emph{et~al.}, ``Vlmevalkit: An open-source toolkit for evaluating large multi-modality models,'' in \emph{Proceedings of the 32nd ACM international conference on multimedia}, 2024, pp. 11\,198--11\,201.

\bibitem{llama2}
H.~Touvron, L.~Martin, K.~Stone, P.~Albert, A.~Almahairi, Y.~Babaei, N.~Bashlykov, S.~Batra, P.~Bhargava, S.~Bhosale \emph{et~al.}, ``Llama 2: Open foundation and fine-tuned chat models,'' \emph{arXiv preprint arXiv:2307.09288}, 2023.

\bibitem{llama3}
A.~Grattafiori, A.~Dubey, A.~Jauhri, A.~Pandey, A.~Kadian, A.~Al-Dahle, A.~Letman, A.~Mathur, A.~Schelten, A.~Vaughan \emph{et~al.}, ``The llama 3 herd of models,'' \emph{arXiv preprint arXiv:2407.21783}, 2024.

\bibitem{wikitext}
S.~Merity, C.~Xiong, J.~Bradbury, and R.~Socher, ``Pointer sentinel mixture models,'' \emph{arXiv preprint arXiv:1609.07843}, 2016.

\bibitem{mmlu}
D.~Hendrycks, C.~Burns, S.~Basart, A.~Zou, M.~Mazeika, D.~Song, and J.~Steinhardt, ``Measuring massive multitask language understanding,'' \emph{arXiv preprint arXiv:2009.03300}, 2020.

\bibitem{hellaswag}
R.~Zellers, A.~Holtzman, Y.~Bisk, A.~Farhadi, and Y.~Choi, ``Hellaswag: Can a machine really finish your sentence?'' \emph{arXiv preprint arXiv:1905.07830}, 2019.

\bibitem{lambada}
A.~Radford, J.~Wu, R.~Child, D.~Luan, D.~Amodei, I.~Sutskever \emph{et~al.}, ``Language models are unsupervised multitask learners,'' \emph{OpenAI blog}, vol.~1, no.~8, p.~9, 2019.

\bibitem{boolq}
C.~Clark, K.~Lee, M.-W. Chang, T.~Kwiatkowski, M.~Collins, and K.~Toutanova, ``Boolq: Exploring the surprising difficulty of natural yes/no questions,'' \emph{arXiv preprint arXiv:1905.10044}, 2019.

\bibitem{arceasy}
P.~Clark, I.~Cowhey, O.~Etzioni, T.~Khot, A.~Sabharwal, C.~Schoenick, and O.~Tafjord, ``Think you have solved question answering? try arc, the ai2 reasoning challenge,'' \emph{arXiv preprint arXiv:1803.05457}, 2018.

\bibitem{piqa}
Y.~Bisk, R.~Zellers, R.~Le~Bras, J.~Gao, and Y.~Choi, ``Reasoning about physical commonsense in natural language,'' 2019.

\bibitem{winogrande}
K.~Sakaguchi, R.~Le~Bras, C.~Bhagavatula, and Y.~Choi, ``An adversarial winograd schema challenge at scale,'' \emph{arXiv preprint arXiv:1907.10641}, 2019.

\bibitem{openbookqa}
T.~Mihaylov, P.~Clark, T.~Khot, and A.~Sabharwal, ``Can a suit of armor conduct electricity? a new dataset for open book question answering,'' \emph{arXiv preprint arXiv:1809.02789}, 2018.

\bibitem{eval-harness}
\BIBentryALTinterwordspacing
L.~Gao, J.~Tow, B.~Abbasi, S.~Biderman, S.~Black, A.~DiPofi, C.~Foster, L.~Golding, J.~Hsu, A.~Le~Noac'h, H.~Li, K.~McDonell, N.~Muennighoff, C.~Ociepa, J.~Phang, L.~Reynolds, H.~Schoelkopf, A.~Skowron, L.~Sutawika, E.~Tang, A.~Thite, B.~Wang, K.~Wang, and A.~Zou, ``A framework for few-shot language model evaluation,'' 12 2023. [Online]. Available: \url{https://zenodo.org/records/10256836}
\BIBentrySTDinterwordspacing

\bibitem{webqs}
\BIBentryALTinterwordspacing
J.~Berant, A.~Chou, R.~Frostig, and P.~Liang, ``Semantic parsing on {F}reebase from question-answer pairs,'' in \emph{Proceedings of the 2013 Conference on Empirical Methods in Natural Language Processing}.\hskip 1em plus 0.5em minus 0.4em\relax Seattle, Washington, USA: Association for Computational Linguistics, Oct. 2013, pp. 1533--1544. [Online]. Available: \url{https://aclanthology.org/D13-1160}
\BIBentrySTDinterwordspacing

\bibitem{gao2020pile800gbdatasetdiverse}
\BIBentryALTinterwordspacing
L.~Gao, S.~Biderman, S.~Black, L.~Golding, T.~Hoppe, C.~Foster, J.~Phang, H.~He, A.~Thite, N.~Nabeshima, S.~Presser, and C.~Leahy, ``The pile: An 800gb dataset of diverse text for language modeling,'' 2020. [Online]. Available: \url{https://arxiv.org/abs/2101.00027}
\BIBentrySTDinterwordspacing

\end{thebibliography}
}

\clearpage
\appendix
\renewcommand{\thesection}{\Alph{section}}  
\twocolumn[
  \centering
  \section*{Appendix}
  \vspace{1em}
]

\section{Group Aware Reordering (GAR) - Extended Description}
\label{sec:Sup-Gar}

In this section, we extend the description of the GAR method.
We consider the W4A16 scheme, i.e., when weights are quantized and stored in 4 bits, and computation is performed in BF16. The same rationale also applies to the W4A8 scheme discussed in this paper.
We begin by revisiting the quantization and dequantization process under the GPTQ algorithm when using per-group scaling and zero-point. 
We then explain the rationale behind activation reordering, followed by a discussion on why unconstrained reordering introduces inference-time overhead. Finally, we present the proposed GAR method and explain how it improves accuracy without incurring any inference overhead.

\subsection{Scales and Zero-Points Computation}
We start by describing the computation of per-group scales and zero-points in the GPTQ quantization algorithm \cite{gptq}.
GPTQ operates iteratively, as described in \cref{ssec:compensating}.
While GPTQ is applied independently to all weights in the column, we describe it for some row $i$ for clarity. The algorithm begins by computing the per-group scale and zero-point, $s^{i,1}$ and $z^{i,1}$, for converting weights of the  first group from BF16 to INT4. These values are computed once at the beginning of the group quantization phase, using \cref{eq:int_quantization}.

\begin{figure}[t]
    \centering
    \includegraphics[width=1\linewidth]{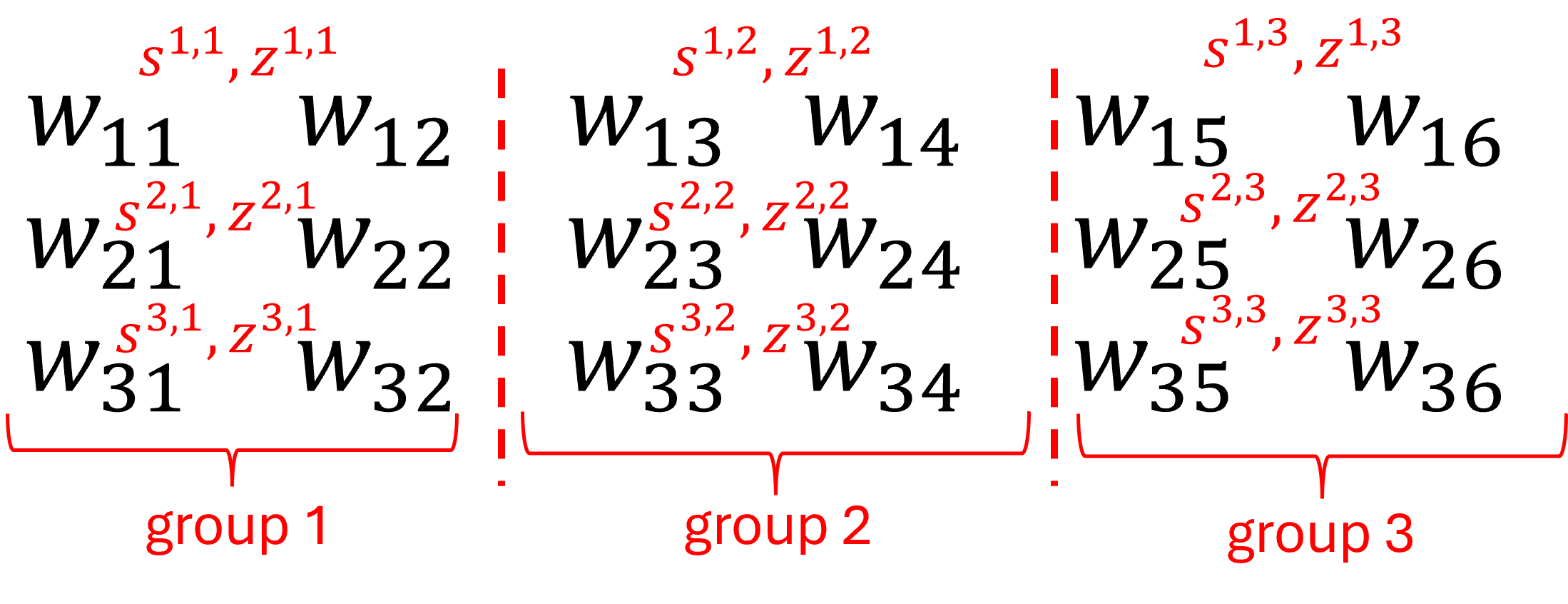}
    \caption{An illustration of per-group scale and zero-point. In this example, the weight tensor is divided into three groups, each containing two weights (i.e., a group size of 2).}
    \label{fig:group_quantization}
\end{figure}

Once $s^{i,1}$ and $z^{i,1}$ are computed, the first weight in the group, $w^{i,1}$ is quantized to INT4 and then dequantized to BF16 using \cref{eq:group_deqauntization}. The quantization error with respect to the original BF16 weight is then calculated and distributed to the remaining unprocessed weights in the same row (i.e., from left to right), following \cref{algorithm_update}.
This process is then repeated for the second group, with a new set of parameters $s^{i,2}, z^{i,2}$, and so on for each subsequent group. At the end of the quantization process, each group of weights has its own scale and zero-point, as shown in \cref{fig:group_quantization}, and the tensor is stored in its 4-bit representation, denoted by $W_4$.

At inference time, the stored 4-bit integer weights are loaded, and each group is dequantized to its 16-bit representation using the formula: 
\begin{equation}\label{eq:group_deqauntization}
\hat{W}^{i,g}_{16} = (W_4^{i,g} - z^{i,g}) \cdot s^{i,g}.    
\end{equation}
Since each scale and zero-point corresponds to a group of \textit{consecutive} weights in the weight tensor, the dequantization for an entire group can be efficiently performed using a single multiplication and subtraction operation, leveraging vectorized computation.

\subsection{Activation Reordering}
We next describe the rationale behind activation reordering, its effect on the scaling factors and zero-points, and why this scheme results in inference-time overhead.

The rationale for activation reordering can be explained as follows.
When using an iterative error compensation method, as described above and implemented in \cite{gptq} and in this paper, it is beneficial to begin by quantizing the most "important" weights. This is because the quantization error from the weights at the beginning of the row is propagated to later weights, which reduces the error for the earlier-positioned weights. Furthermore, the weights that appear at the end of the row accumulate more error due to previous updates, resulting in higher quantization error. Thus, weights that are quantized earlier will typically incur less quantization error.

To determine which weights are the most "important", we consider the diagonal of the Hessian matrix. These diagonal elements reflect the relative significance of different input features, as they correspond to the second-order sensitivity of the loss with respect to each weight, as explained intuitively next. Recall that for a linear layer with input $X$, the Hessian under our objective in \cref{The GPTQ_algorithm} is given by $X^T X$. In this case, the diagonal elements of the Hessian reflect the squared magnitudes of the input features. Consequently, weights associated with larger diagonal values contribute more significantly to the output. Modifying such weights has a stronger effect on the model’s behavior, making them more critical to quantize accurately.

Combining the above observations that (1) earlier-quantized weights incur lower error and (2) the most important weights are those associated with the largest Hessian diagonal entries, it is advantageous to reorder the weights with respect to the Hessian diagonal before the iterative quantization process. Specifically, this is done by permuting the Hessian in descending order with respect to its diagonal elements, and then reordering the weight tensor accordingly. This is precisely the activation reordering technique proposed in \cite{gptq}.
Empirically, this technique has been shown to enhance the accuracy of quantized models and is widely adopted in practice. 

However, despite its accuracy benefits, the reordering technique introduces additional overhead during inference (i.e., increased latency) as discussed next.

When applying the activation reordering technique, we first permute the weight tensor and then run the iterative algorithm as described at the beginning of this section. Specifically, the scale factors and zero-points are computed for groups of the \textit{permuted} weight tensor. At the end of the process, the weight tensor is re-permuted to its original order to ensure correct multiplication during inference.

\begin{figure}[t]
    \centering
    \includegraphics[width=1\linewidth]{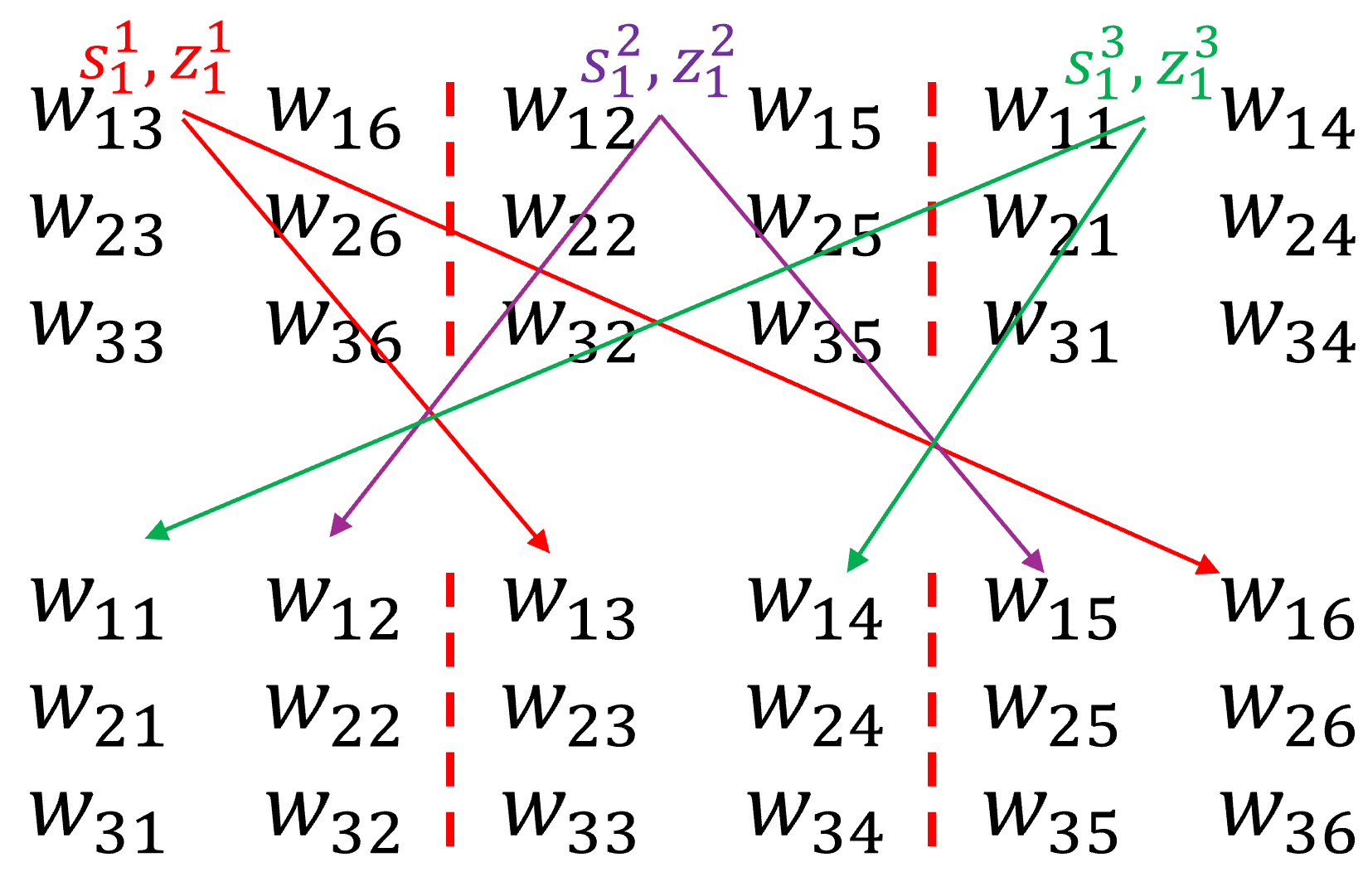}
    \caption{Computing scales and zero-points under full activation reordering scheme. 
    As can be seen in this example in the top matrix, the third and sixth columns have been reordered to the first and second columns. Thus during inference, their scale and zero point need to be fetched from these groups, which are not consecutive in memory, and require indexing.}
    \label{fig:act_order}
\end{figure}

However, after re-permutation, the scale factors and zero-points are no longer aligned with the weight groups. That is, each weight in the tensor may now have a different scale and zero-point, as illustrated in \cref{fig:act_order}. As a result, the dequantization in \cref{eq:group_deqauntization} can no longer be efficiently applied using a single scalar subtraction and multiplication per group. Instead, each weight must be individually processed with its corresponding zero-point and scale factor, increasing the number of operations required during inference and thereby increasing latency.

\subsection{Group Aware Reordering}
\begin{figure}[t]
    \centering
    \includegraphics[width=1\linewidth]{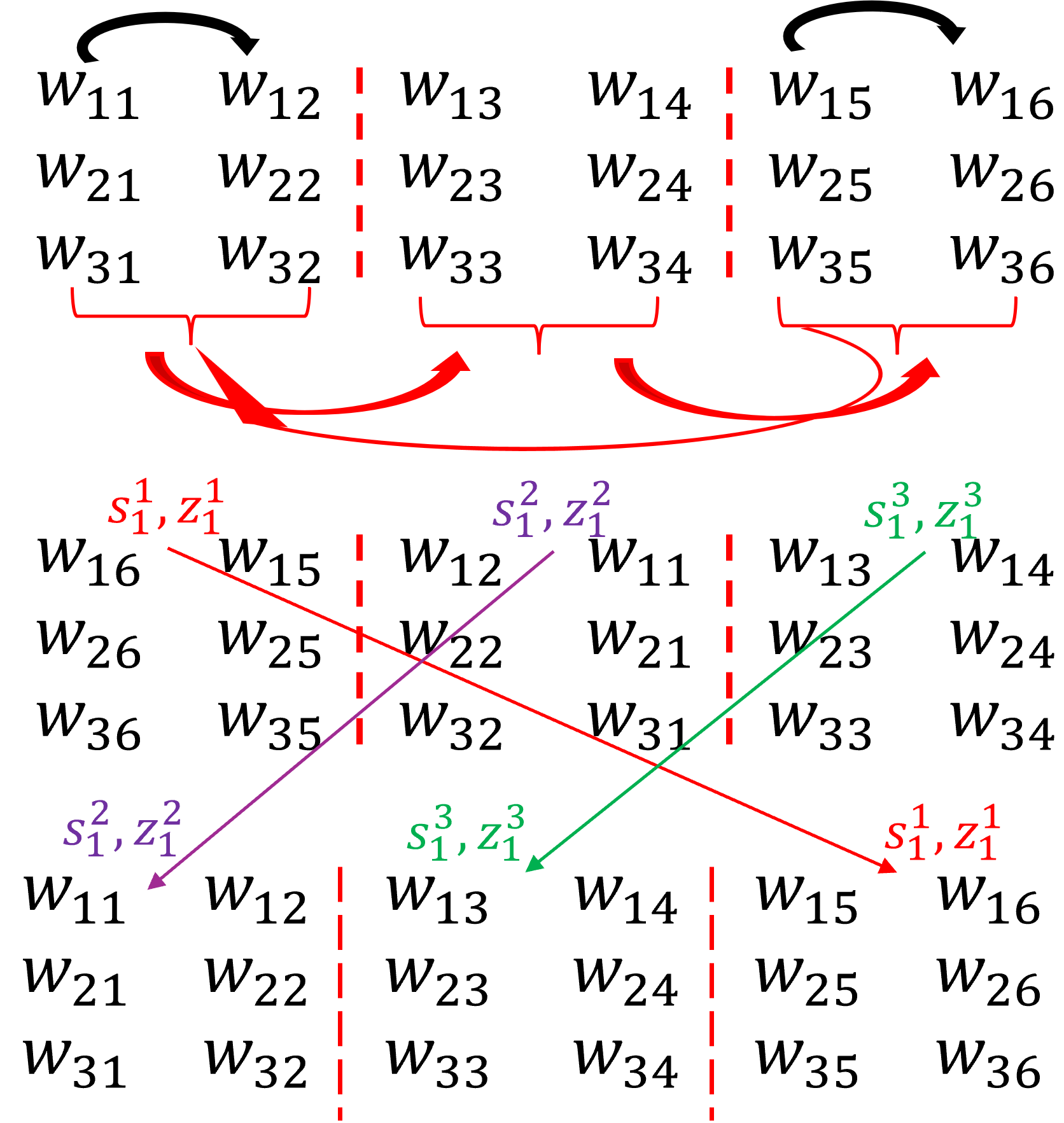}
    \caption{Computing scales and zero-points under GAR method.}
    \label{fig:gar_illustrated}
\end{figure}

The proposed GAR method addresses the issue of inference overhead while still allowing the weight tensor to be reordered based on weight importance, subject to certain constraints. Specifically, as described in \cref{ssec:GAR}, GAR allows permutations in two ways: (i) within a group, and/or (ii) by rearranging entire groups, as illustrated in the top panel of \cref{fig:gar_illustrated}.

This constrained permutation ensures that, after computing the scales and zero-points for the permuted tensor, each original group of weights shares the same scale and zero-point, as shown in the middle panel of \cref{fig:gar_illustrated}. As a result, when the weight tensor is re-permuted back to its original order, the associated scales and zero-points can be re-permuted accordingly (see the bottom panel of \cref{fig:gar_illustrated}). This guarantees that each group of weights retains its own scale and zero-point, allowing \cref{eq:group_deqauntization} to be efficiently implemented during inference, as in the original scheme without activation reordering.

Although our method imposes constraints on the activation order, prior studies have shown that most layers are dominated by a small number of highly important output activations \cite{super, massive}. Therefore, even though our scheme only allows a partial ordering of activations, it effectively preserves the most significant weights. As a result, we expect only a small accuracy gap compared to the full reordering scheme.
This hypothesis is supported by our results in \Cref{tab:ablation}. Under both W4A16 and W4A8 quantization settings, the GAR method delivers a significant accuracy improvement over the baseline without reordering, while incurring only a minor accuracy drop relative to the full reordering approach.

Finally, we note that GAR requires a ranking criterion to determine which group should be quantized first. In our experiments, we ranked groups based on the maximum value of the Hessian diagonal within each group, that is, we compared the maximum diagonal entry of each group of weights. Alternative criteria are possible, for example, ranking groups by the average of the top 10\% largest elements before sorting. We leave the exploration of other ranking strategies for future work.


\end{document}